\documentclass[preprints,article,accept,pdftex,moreauthors]{Definitions/mdpi} 
\firstpage{1} 
\pubvolume{1}
\issuenum{1}
\articlenumber{0}
\pubyear{2025}
\copyrightyear{2025}
\datereceived{ } 
\daterevised{ } 
\dateaccepted{ } 
\datepublished{ } 
\hreflink{https://doi.org/} 

\Title{High-Definition 5MP Stereo Vision Sensing for Robotics}

\TitleCitation{Title}

\Author{Leaf Jiang $^{1,}$*\orcidA{}, Matthew Holzel $^{2}$, Bernhard Kaplan $^{2}$, Hsiou-Yuan Liu $^{1}$, Sabyasachi Paul $^{2}$, Karen Rankin $^{1}$, and Piotr Swierczynski $^{2}$\orcidB{}}

\AuthorNames{Leaf Jiang, Matthew Holzel, Bernhard Kaplan, Hsiou-Yuan Liu, Sabyasachi Paul, Karen Rankin, and Piotr Swierczynski}

\isAPAStyle{%
       \AuthorCitation{Lastname, F., Lastname, F., \& Lastname, F.}
         }{%
        \isChicagoStyle{%
        \AuthorCitation{Lastname, Firstname, Firstname Lastname, and Firstname Lastname.}
        }{
        \AuthorCitation{Lastname, F.; Lastname, F.; Lastname, F.}
        }
}

\address{%
$^{1}$ \quad NODAR Inc.; 240 Oak Hill Cir, Concord, MA 01742, USA\\
$^{2}$ \quad NODAR Sensor GmbH; Sirius Business Park Berlin-Tempelhof, Großbeerenstraße 2-10, Geb 1, EG, Raum 13/14, 12107 Berlin}

\corres{Correspondence: leaf@nodarsensor.com; Tel.: +01-855-218-6422}

\abstract{High-resolution (5MP+) stereo vision systems are essential for advancing robotic capabilities, enabling operation over longer ranges and generating significantly denser and accurate 3D point clouds. However, realizing the full potential of high-angular-resolution sensors requires a commensurately higher level of calibration accuracy and faster processing -- requirements often unmet by conventional methods. This study addresses that critical gap by processing 5MP camera imagery using a novel, advanced frame-to-frame calibration and stereo matching methodology designed to achieve both high accuracy and speed. Furthermore, we introduce a new approach to evaluate real-time performance by comparing real-time disparity maps with ground-truth disparity maps derived from more computationally intensive stereo matching algorithms. Crucially, the research demonstrates that high-pixel-count cameras yield high-quality point clouds only through the implementation of high-accuracy calibration.}

\begin{document}




\section{Introduction}

High-fidelity three-dimensional (3D) environmental sensing is a fundamental requirement for advanced robotic and autonomous systems. Stereo vision, in particular, offers a passive, cost-effective, and scalable method for generating dense 3D representations, or point clouds, crucial for tasks like path planning, object manipulation, and long-range localization. Recent technological advancements have led to the widespread adoption of high-resolution sensors, such as 5-megapixel (5MP) stereo camera systems. These high-angular-resolution sensors are essential for next-generation robotic capabilities, offering the promise of operating at significantly longer ranges, integrating with higher-capacity payloads, and providing a dramatic increase in the density and accuracy of the resulting point clouds.

The necessity of high-resolution sensing extends across the operational spectrum, from macro-scale scene analysis to micro-scale manipulation. At long ranges, high resolution is critical for reliably detecting and identifying small objects, preventing them from being blended into the background (akin to a low-pass filtering effect) and ensuring high confidence in tracking and recognition. Conversely, at close range, the sub-millimeter precision afforded by these systems is crucial for fine-motor tasks, such as the robotic manipulation and inspection of delicate components, including specialized parts like dental implants or micro-assembly units. This dual requirement highlights the increasing demand for robust, high-performance stereo vision systems.

The transition to high-resolution sensors, however, introduces a critical dependency: the precision required for camera system calibration must increase proportionally to the sensor's horizontal pixel count. Specifically, the ability to resolve fine details and accurately determine depth (range resolution) at increasing distances relies not just on the raw sensor capability but also on the sub-pixel accuracy of the entire stereo rig model. Conventional static stereo camera calibration methodologies \cite{opencv_library}, involving a one-time calibration at the factory, which are typically sufficient for lower-resolution systems, often introduce residual errors that become dominant in 5MP and higher-resolution setups over time, especially for systems exposed to shock and vibration, such as walking robots or autonomous vehicles. These limiting errors prevent the realization of the theoretical point cloud quality that the hardware is capable of achieving, creating a significant performance gap in state-of-the-art robotic vision systems. Instead, dynamic calibration \cite{nodar_patent}, which can be calibrated on natural scenes, is necessary to maintain the performance of high-definition stereo vision systems under changing environmental conditions and potentially harsh conditions.

To contextualize the commercial landscape driving this research, Table~\ref{tbl:stereo_cameras} summarizes the characteristics of several modern commercial stereo camera products and their respective resolutions. With higher-resolution cameras and longer baselines (the distance between cameras), extrinsic camera calibration becomes more difficult due to greater angular tolerance and greater sensitivity to disturbances. For example, the bending angle (slope of deflection) of a cantilever beam (representing the stereo camera structure) with a point load applied at the free end is proportional to the \emph{square} of the length of the beam. This means that a 1-m baseline stereo camera is 100 times more sensitive to external perturbative forces arising from shocks and vibrations than a 10-cm baseline stereo camera, necessitating faster, more accurate online extrinsic camera calibration.

Table~\ref{tbl:stereo_cameras} highlights a second trend towards outdoor robotics, where high-dynamic-range (HDR) operation is crucial, enabling the stereo camera to obtain depth estimates in challenging lighting conditions, such as deep shadows or when the sun is shining directly into the camera. Cameras that offer this crucial high dynamic range (typically, 120-150 dB for automotive CMOS) often employ a rolling shutter rather than the preferred global shutter. This rolling-shutter mechanism inherently introduces difficulties for standard stereo matching algorithms. However, this barrier is being overcome: a few manufacturers, such as NODAR, are now producing specialized stereo vision processing solutions that successfully work with rolling-shutter HDR cameras.

The ``Min Light" column in Table~\ref{tbl:stereo_cameras} is, more accurately, the absolute sensitivity threshold, which is the minimum number of photons needed to equal the noise level. The NODAR stereo camera, for example, has a nearly photon-counting sensitivity with an absolute sensitivity threshold of 1.5~photons, which allows this camera to operate at night with excellent sensitivity.

\begin{table}[H]
\small 
\caption{Summary of modern commercial stereo vision camera system resolutions.}
\label{tbl:stereo_cameras}
\isPreprints{\centering}{} 
\begin{tabularx}{\textwidth}{l c c c c c}
\toprule
\textbf{Product Model} & \textbf{Manufacturer} & \textbf{Resolution} & \textbf{Dynamic Range} & \textbf{Min Light} & \textbf{Baseline} \\
\midrule
NDR-HDK-2.0-100 & NODAR & \shortstack{5.4MP \\ $2880 \times 1860$} & 123 dB & 1.5 photons & 100 cm \\
\\
Eagle & \shortstack{Leopard\\Imaging} & \shortstack{5.1 MP \\$2560 \times 1984$} & 100 dB & -- & 15 cm \\
\\
ZED X & Stereolabs & \shortstack{2.3 MP \\ $1920 \times 1200$} & 71.4 dB & -- & 12 cm \\
\\
Bumblebee X & \shortstack{Teledyne\\FLIR} & \shortstack{3.2 MP \\ $2048 \times 1536$} & 71.62 dB & 4.64 photons & 24 cm \\
\\
D455 & Realsense & \shortstack{0.92 MP \\ $1280 \times 720$} & -- & -- & 9.5 cm \\
\\
OAK 4 D & Luxonis & \shortstack{1 MP \\ $1280 \times 800$} & 68 dB & -- &  7.5 cm \\
\\
Gemini 435Le & Orbbec & \shortstack{1 MP \\ $1280 \times 800$} & -- & -- & 9.5 cm\\
\bottomrule
\end{tabularx}
\end{table}

This study directly addresses the performance bottleneck imposed by insufficient calibration. We propose and validate a novel, advanced calibration methodology specifically engineered to meet the demanding accuracy requirements of 5MP stereo vision. By meticulously processing imagery from two 5MP cameras using this advanced methodology, we successfully demonstrate the realization of the system's theoretical point cloud quality. A key finding of this work is the establishment of a novel scaling law: point cloud quality, defined as the range resolution, scales as the square-root of the number of camera pixels. Our results show that high pixel count alone is insufficient; high-quality point clouds are achieved only with this specialized advanced calibration. The remainder of this paper is structured as follows: Section 2 reviews theoretical framework. Section 3 details the materials and methods. Section 4 presents the experimental results. Finally, Section 5 provides the conclusion.

\section{Theoretical Framework}

This section discusses the online calibration algorithms and derives the equation that relates the point cloud quality to the image sensor resolution.

\subsection{Online Auto-Calibration Algorithm}
\begin{figure}
    \centering
    \includegraphics[width=0.75\linewidth]{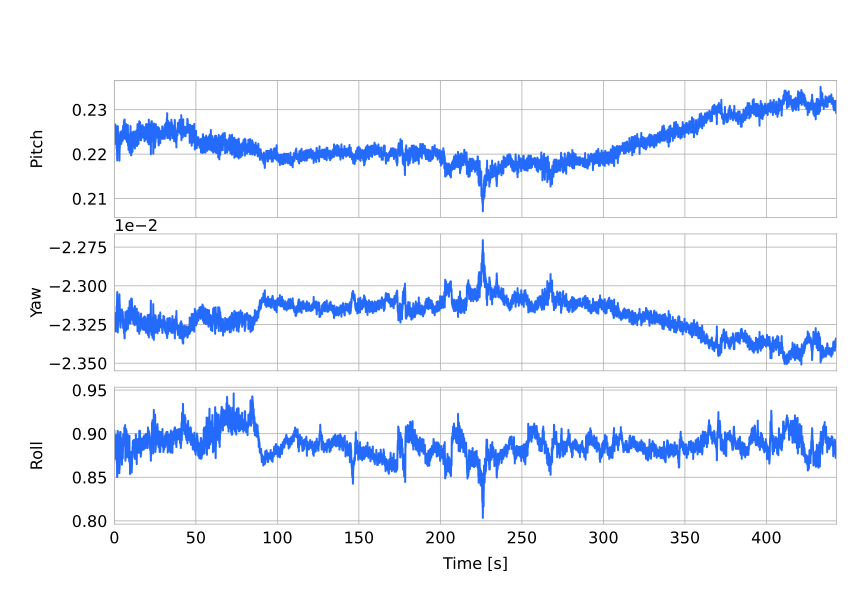}
    \caption{Extrinsic camera parameters change every frame for 1.2-m baseline stereo camera mounted on a car and driving on a highway. The relative roll, pitch, and yaw are shown in units of degrees.}
    \label{vibration}
\end{figure}

Auto-calibration of a stereo rig is discussed in \cite{hartley}, but such approaches rely on keypoints, which are inaccurate in natural scenes with non-pointy, round structures. These algorithms cannot be computed quickly enough to compensate for perturbations that occur frame by frame (e.g., engine or road vibration). For example, the relative roll, pitch, and yaw between two cameras mounted 1.2 meters apart on a car driving on a highway are plotted in Fig.~\ref{vibration}, showing that the extrinsic camera parameters vary significantly from frame to frame. These extrinsic camera parameters are extracted from the images and do not ues any accelerometer or inertial sensor data, which makes the approach more straightforward to implement, as it only needs camera images.

The latest auto-calibration algorithms, such as \cite{nodar_patent}, do not make keypoint assumptions and consider all pixels in the frame (not just a few select keypoints that are often clustered in small parts of the image with unique texture), which are sensitive and fast enough to reveal the high-frequency camera motion in Fig.~\ref{vibration}. In this paper, we use the NODAR auto-calibration algorithm, downloaded from \cite{hammerhead_sdk}.

\subsection{Stereo Matcher}


Selecting a stereo matching algorithm involves a trade-off between computational efficiency (speed) and disparity map accuracy. This research investigates two distinct stereo matching approaches developed by NODAR, each optimized for a specific set of robotic application requirements. The computational benchmarks are shown in Table~\ref{tbl:benchmarks}. It is important to note that the algorithms operate at native resolution and are not downsampled, as downsampling can miss small objects or those far away ($\sim 5-15$~pixels).

\begin{table}[H]
\small 
\caption{Computational benchmarks for Hammerhead and GroundTruth stereo matching algorithms on popular embedded GPUs and a common gaming GPU. The reported resolutions are depth maps at their native resolution (not downsampled, as downsampling is a common first step for stereo matcher neural networks to increase frame rates). This allows measuring the distance to small objects since it is at the native resolution. The specification, Points Per Second (pps), is common for lidar systems.}
\label{tbl:benchmarks}
\isPreprints{\centering}{} 
\begin{tabularx}{\textwidth}{l c c c}
\toprule
\textbf{Hardware} & GPU & \textbf{Hammerhead} & \textbf{GroundTruth} \\
\midrule
Nvidia AGX Orin & 5.3 FP32 TFLOPS & \shortstack{160M pps \\ 20 FPS @ 8MP \\ 30 FPS @ 5.4MP \\ 160 FPS @ 1MP} & \shortstack{10M pps \\ 1 FPS @ 8MP \\ 2 FPS @ 5.4MP \\ 10 FPS @ 1MP} \\
\\
Nvidia AGX Thor & 7.8 FP32 TFLOPS & \shortstack{235M pps \\ 29 FPS @ 8MP \\ 44 FPS @ 5.4MP \\ 235 FPS @ 1MP} & \shortstack{15M pps \\ 1.8 FPS @ 8MP \\ 2.7 FPS @ 5.4MP \\ 15 FPS @ 1MP} \\
\\
Nvidia Thor-X-Super & 18.4 FP32 TFLOPS & \shortstack{555M pps \\ 69 FPS @ 8MP \\ 103 FPS @ 5.4MP \\ 555 FPS @ 1MP} & \shortstack{35M pps \\ 4.3 FPS @ 8MP \\ 6.4 FPS @ 5.4MP \\ 35 FPS @ 1MP} \\
\\
Nvidia RTX 4090 & 82.6 FP32 TFLOPS & \shortstack{2494M pps \\ 312 FPS @ 8 MP \\ 462 FPS @ 5.4MP \\ 2494 FPS @ 1MP} & \shortstack{156M pps \\ 19.5 FPS @ 8MP \\ 28.9 FPS @ 5.4MP \\ 156 FPS @ 1MP} \\
\\
\bottomrule
\end{tabularx}
\end{table}

\subsubsection{Real-Time Stereo Matching: The Hammerhead Algorithm}
The NODAR Hammerhead algorithm \cite{hammerhead} is implemented as a real-time stereo matcher. It is specifically optimized for applications that demand high throughput, such as those in autonomous robotic systems. This algorithm is capable of rapidly processing 5-megapixel (MP) images at their full native resolution, addressing the critical requirement for low-latency operation in dynamic environments. 

\subsubsection{High-Accuracy Stereo Matching: The GroundTruth Algorithm}
In contrast, the NODAR GroundTruth algorithm \cite{groundtruth} is used as an offline stereo matcher. The primary optimization objective of this algorithm is to achieve the highest possible disparity map accuracy, with computational efficiency a secondary consideration. This makes it particularly suitable for applications that are less constrained by real-time processing demands, such as generating precise training data, validating algorithmic performance, and establishing ground-truth references.

The inclusion of these two algorithms effectively spans the primary use-case spectrum encountered in robotics development: from offline training and validation requiring maximum accuracy (GroundTruth) to on-board product implementation necessitating a highly efficient, lightweight, and real-time solution (Hammerhead).

\subsection{Quantifying Depth Uncertainty in Stereo Vision}
The best theoretical depth resolution for a stereo camera, limited by the camera's angular resolving power, is \cite{szeliski}
\begin{linenomath}
\begin{equation}
    \Delta z_{best} = \frac{z^2}{fB} \Delta d = \frac{x^2}{B} \Delta \theta,
\end{equation}
\end{linenomath}
where $z$ is the depth to the object, $f$ is the focal length (typically units of pixels), $B$ is the baseline width, and $\Delta d$ is the disparity resolution (typically in units of pixels). The angular resolution of the stereo vision measurement is $\Delta \theta = \Delta d/f$. Since the angular resolution of the camera is proportional to the square-root of the number of pixels, $N_{pix}$, then the depth resolution can be rewritten as
\begin{linenomath}
\begin{equation}
    \Delta z_{best} \propto \frac{z^2}{B} \sqrt{N_{pix}}.
\end{equation}
\end{linenomath}
Therefore, the depth resolution improves with a longer baseline or a larger format camera. In this paper, we experimentally show that increasing $N_{pix}$ from 1.275 to 5.1 MP (a factor of 4) improves the range resolution (point cloud quality) by a factor of 2, thereby confirming the square-root functional form.


The standard methodology for quantifying depth uncertainty ($\Delta z$) in stereo vision typically involves calculating the root-mean-square error (RMSE) of the disparity measurements around a planar target with a highly textured surface.

\subsubsection{Limitations of the Standard Approach}
While useful for baseline assessment, this conventional method often fails to capture the full spectrum of large, sporadic stereo-matching errors that significantly impact the perceived quality and robustness of depth data in real-world scenarios. The use of an artificially textured, planar target simplifies the matching problem, often leading to an overly optimistic assessment of algorithm performance.

\subsubsection{Proposed Methodology: GroundTruth Approach}
In contrast, our approach utilizes real-world objects of interest, specifically human subjects and complex, non-planar surfaces with challenging geometry and reflectance characteristics, to derive a more representative and robust measure of depth quality. 

Using the high-accuracy stereo matcher disparity map as the ground truth (NODAR GroundTruth), we can then directly compute the error of the actual disparity map generated by the real-time stereo matcher algorithm (NODAR Hammerhead) as a function of range bin. More specifically, the range error for a given depth bin, $\Delta z_n$ from $z=z_n$ to $z_{n+1}$, is the RMS difference between Hammerhead, $z_{hh}$, and GroundTruth, $z_{gt}$, depth estimates for pixels that are in the depth bin.

This methodology provides a more meaningful evaluation compared to relying on an unrealistic, texture-rich calibration target that facilitates trivial matching.

\section{Materials and Methods}





\begin{figure}[H]
\isPreprints{\centering}{} 
\includegraphics[width=0.75\textwidth]{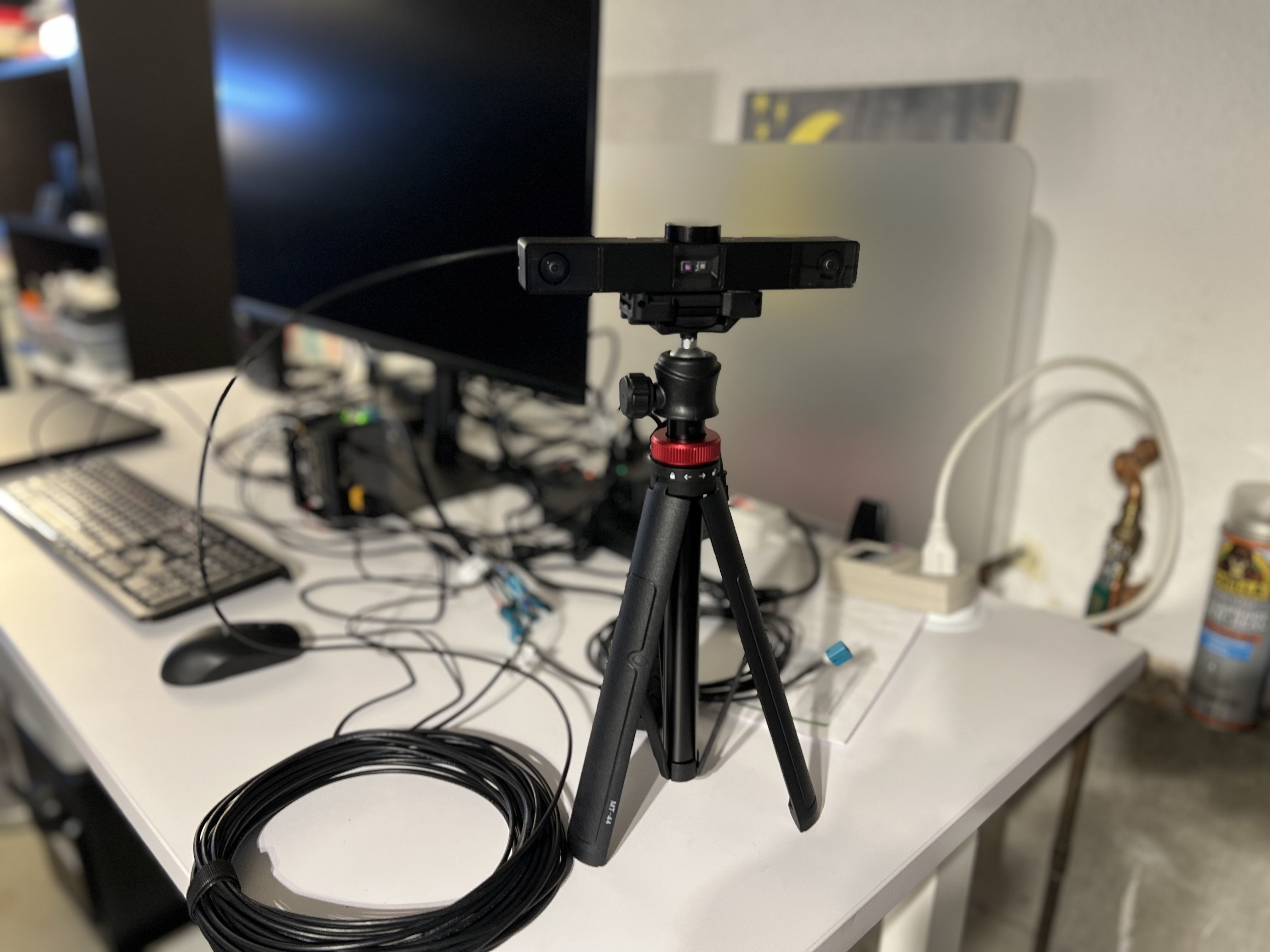}
\caption{The 15-cm-baseline stereo vision camera used in this work.}
\label{leopard_eagle}
\end{figure}

The stereo vision camera used in the experiments, shown in Figure~\ref{leopard_eagle}, is the Leopard Imaging Eagle (model LI-VB1940-GM2C-137H) with a wide 137-degree horizontal field of view \cite{leopard_datasheet}. This system incorporates two 5.1-megapixel, 2560 (H) x 1984 (V), STMicroelectronics VB1940 automotive image sensors. Both sensors feature a global shutter and are of the RGB-IR type. The two cameras are separated by 15 cm and utilize Maxim GMSL2 serializers for data transmission.

The system setup consisted of a stereo camera linked to an Nvidia Jetson Orin AGX developer kit via a Leopard Imaging serializer board. The Orin AGX was loaded with NODAR Hammerhead software, version 2.8.0 \cite{hammerhead_sdk}. We utilized the NODAR Hammerhead camera driver for Leopard Imaging Eagle camera \cite{nodarhub_leopard}. This driver captures the stereo camera feed via GStreamer wrapper and translates it into ZMQ packets for Hammerhead processing. The camera-to-camera timing jitter was measured by recording a stopwatch on a monitor and was found to be about 1 ms.

The system was deployed outdoors on a tripod to capture scenes with long-range depth (relative to the baseline distance or about 200x baseline length) and high dynamic range (i.e., bright and dark areas). During data collection, the following were recorded to disk: full-resolution color images from both left and right cameras, and corresponding depth maps. Some sky pixels were saturated, illustrating the performance of a realistic stereo camera system with suboptimal gain and exposure settings.

In post-processing, the images were resized to half the height and half the width with a Lanczos resampling \cite{magick}.

A single outdoor data capture was processed four ways:
\begin{enumerate}
    \item GroundTruth with 5MP images
    \item GroundTruth with downsampled images, 1.3MP
    \item Hammerhead with 5MP images
    \item Hammerhead with downsampled images, 1.3MP
\end{enumerate}
The scene contained grass, trees, and moving people.

\section{Results}


A single video sequence of 82 frames at approximately 5 FPS was captured from the left and right cameras of a grassy area with people moving around. The stereo camera was mounted on a tripod at about 1 m height. The raw data used in the article, as well as the Google Colab notebook used to summarize the data, are available on a public Google Drive \cite{gdrive_data}.

\begin{figure}[H]
\isPreprints{}{
\begin{adjustwidth}{-\extralength}{0cm}
\centering
} 
\subfloat[\centering]{\includegraphics[width=4.6cm]{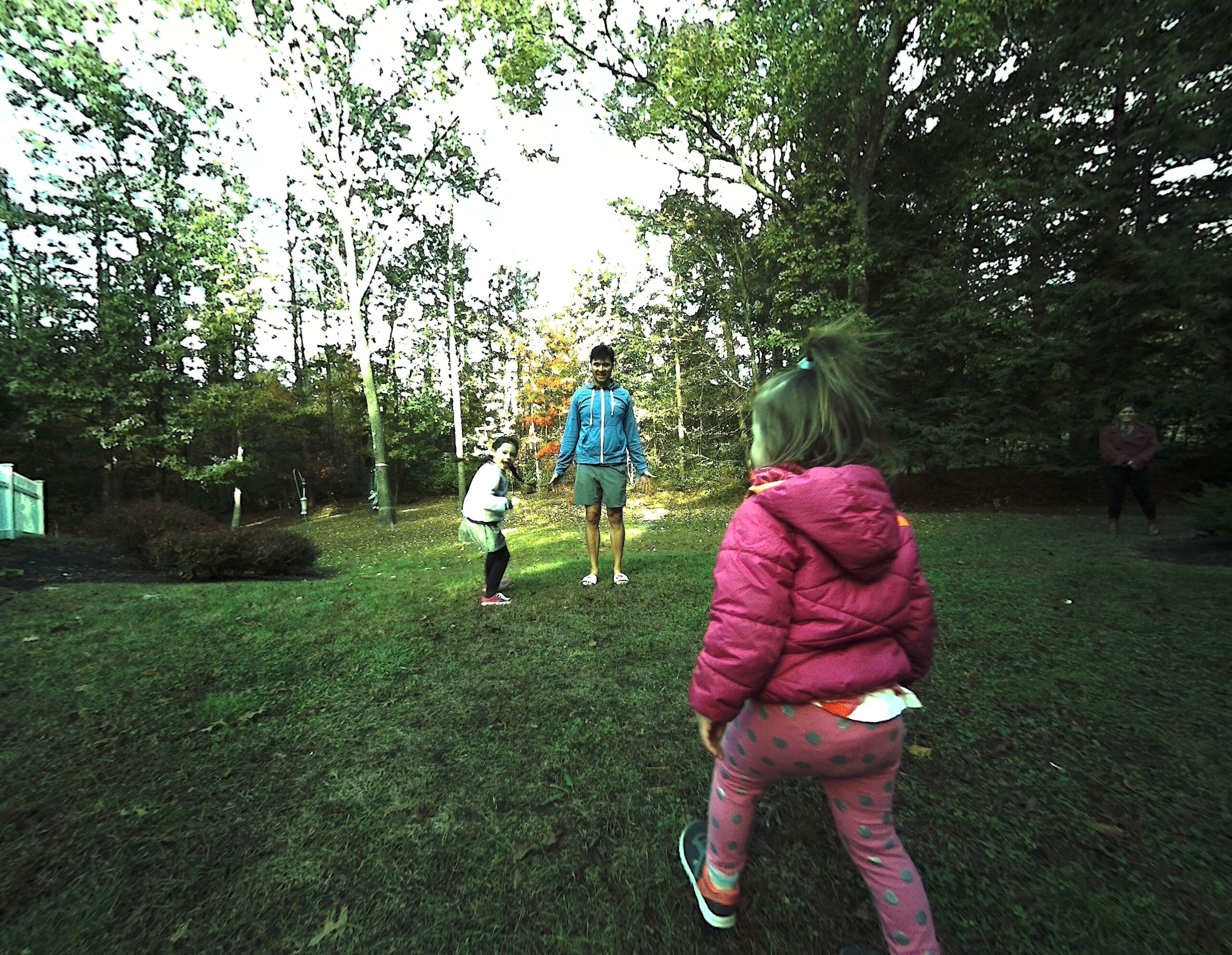}}
\subfloat[\centering]{\includegraphics[width=4.6cm]{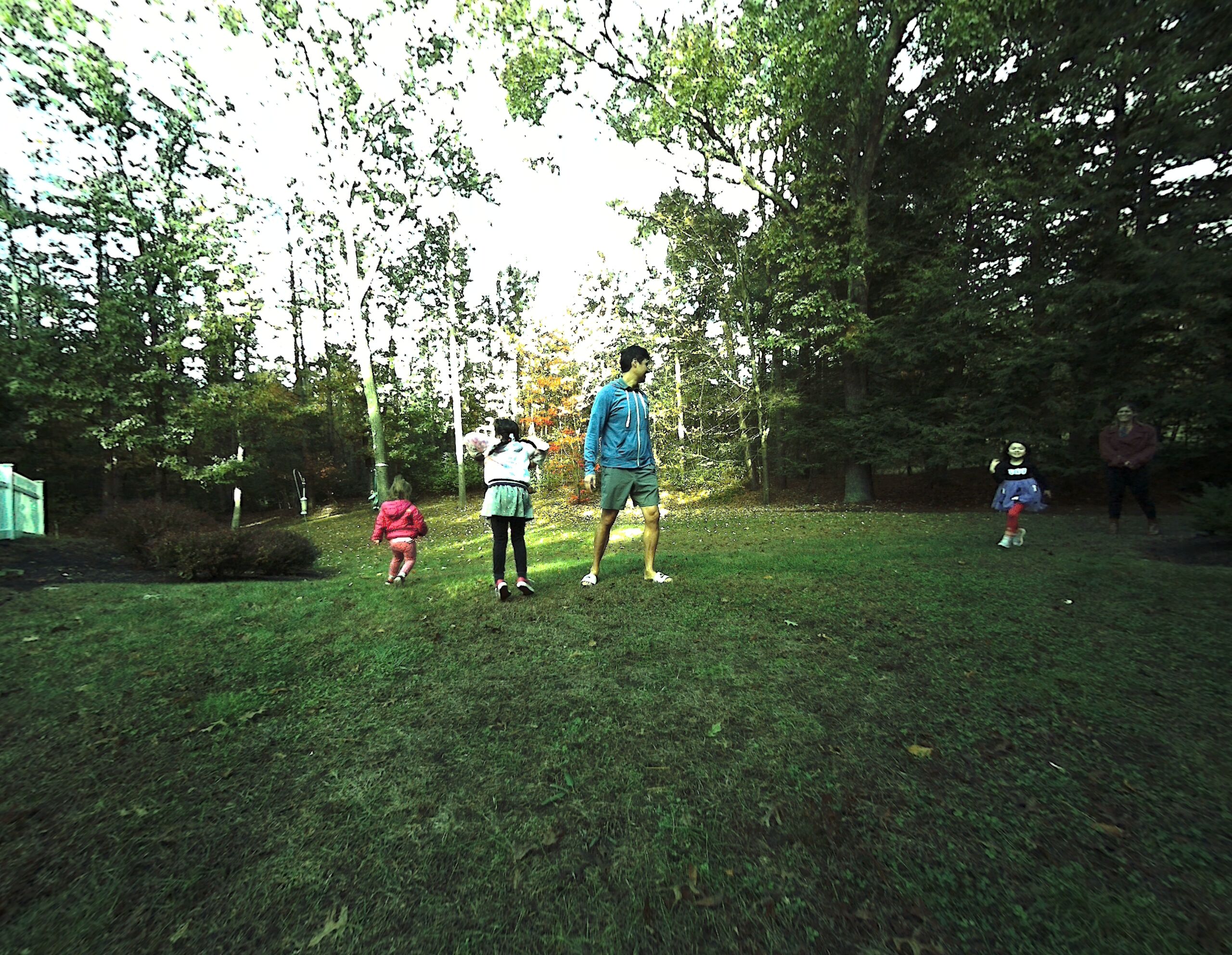}}
\subfloat[\centering]{\includegraphics[width=4.6cm]{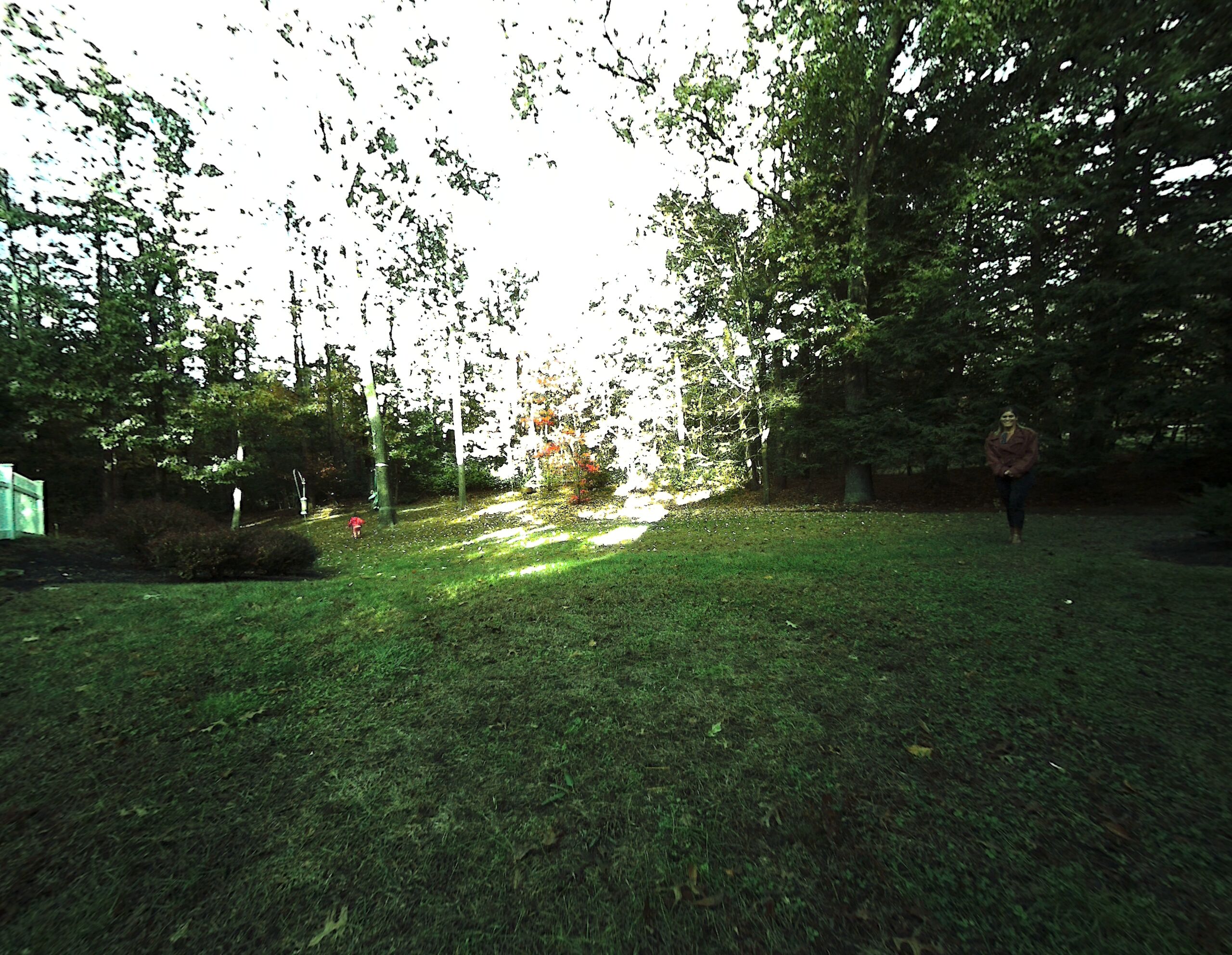}}
\isPreprints{}{
\end{adjustwidth}
} 
\caption{Left rectified images: (\textbf{a}) Frame 10. (\textbf{b}) Frame 17. (\textbf{c}) Frame 41.}
\label{left-rect}
\end{figure}

The left rectified images for frames 10, 17, and 41 are shown in Fig.~\ref{left-rect}, and were used for more detailed analysis. These frames were selected to capture the person in the pink jacket at near (0.71 m), intermediate (4.50 m), and far (20.93 m) ranges.

\begin{figure}[H]
\isPreprints{}{
\begin{adjustwidth}{-\extralength}{0cm}
\centering
} 
\subfloat[\centering]{\includegraphics[width=7.0cm]{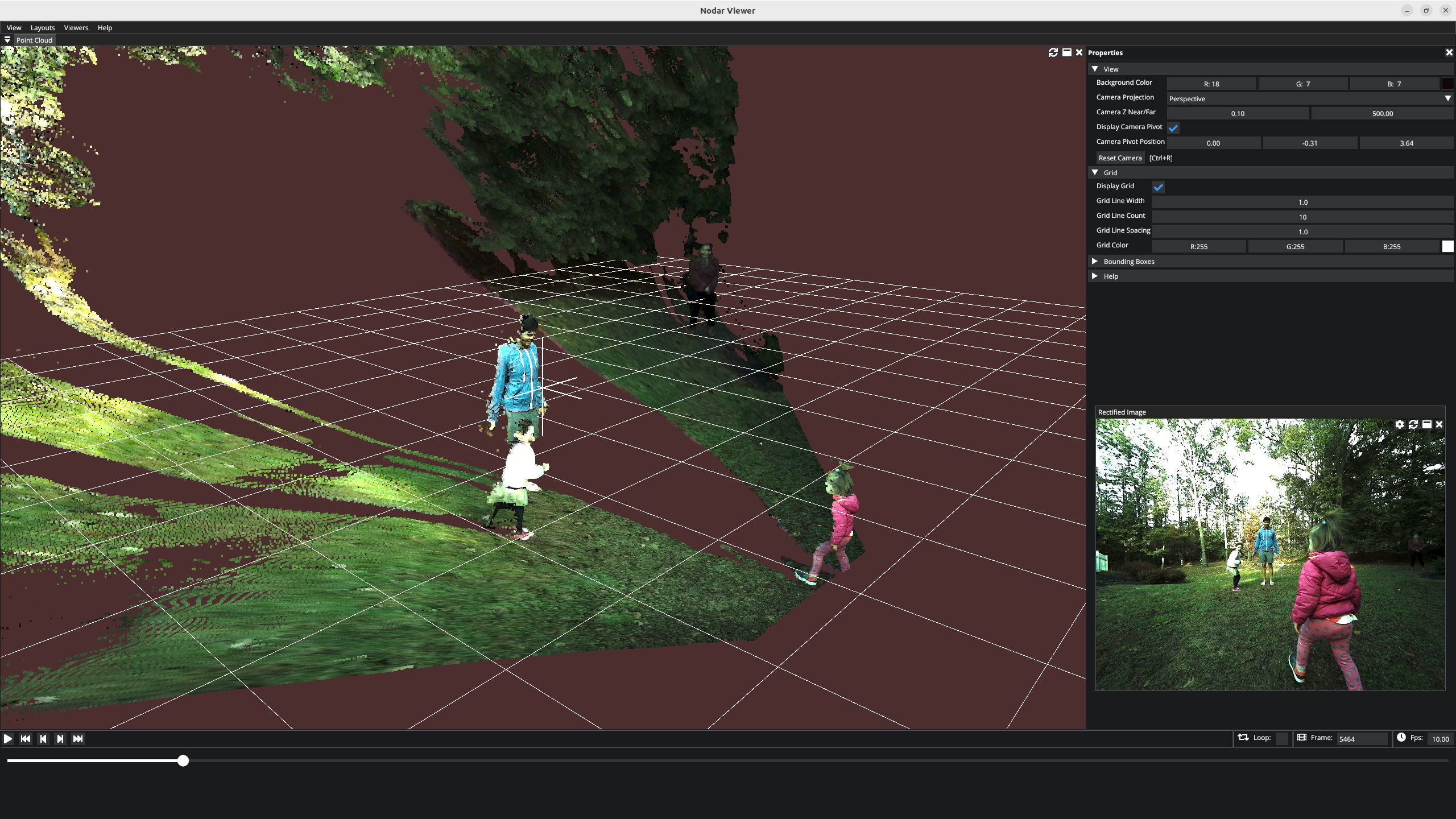}}
\subfloat[\centering]{\includegraphics[width=7.0cm]{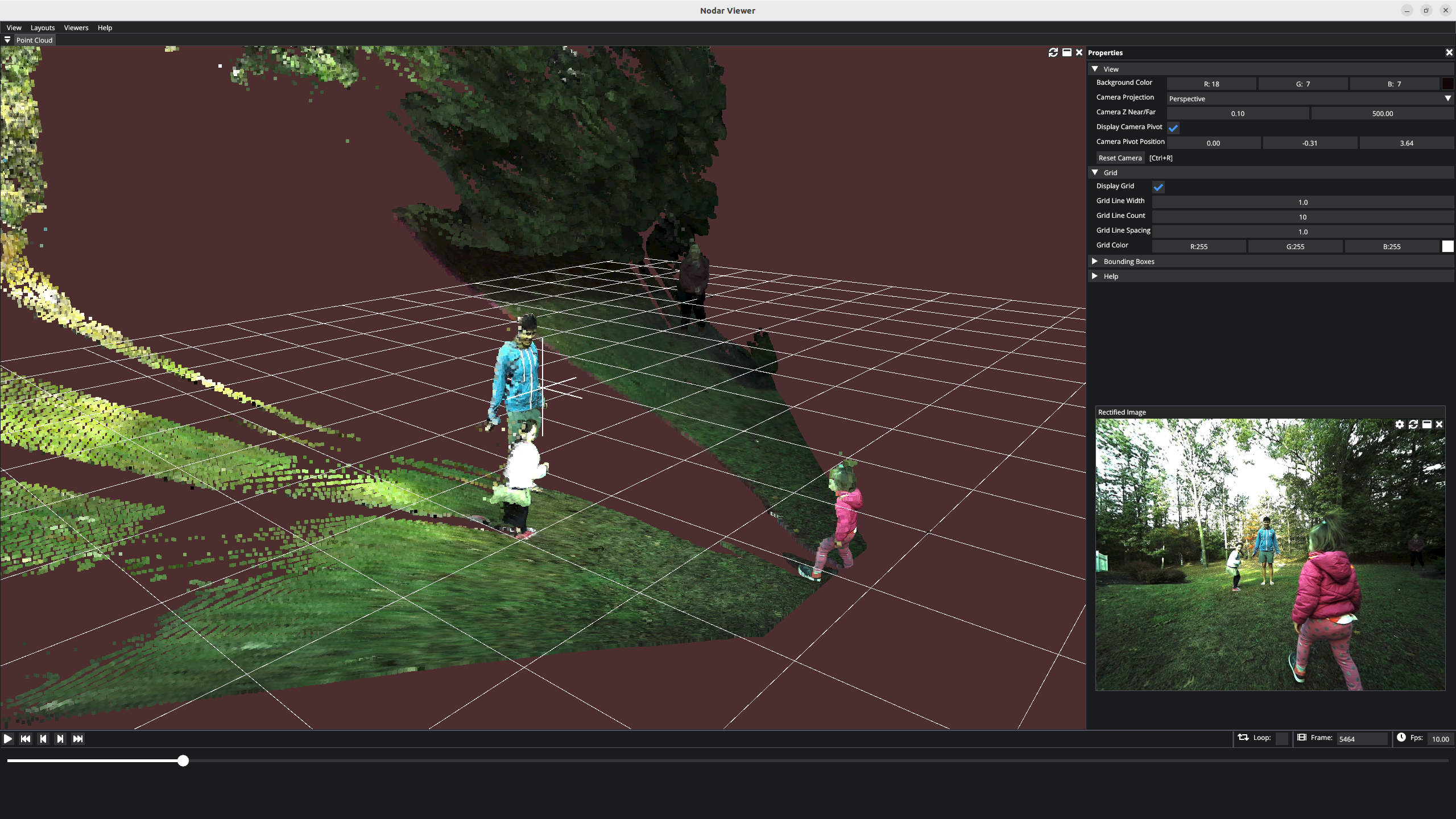}}\\
\subfloat[\centering]{\includegraphics[width=7.0cm]{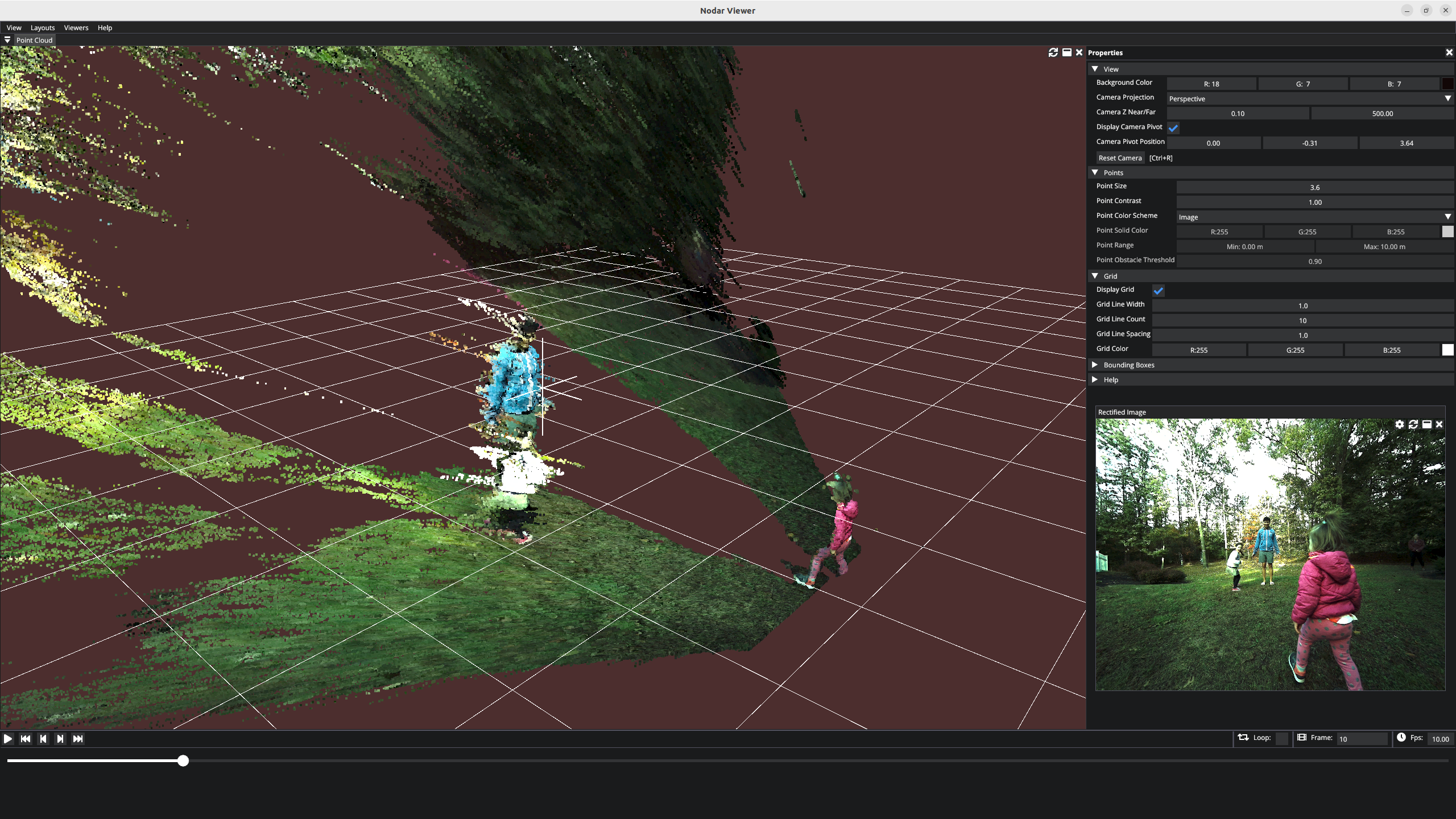}}
\subfloat[\centering]{\includegraphics[width=7.0cm]{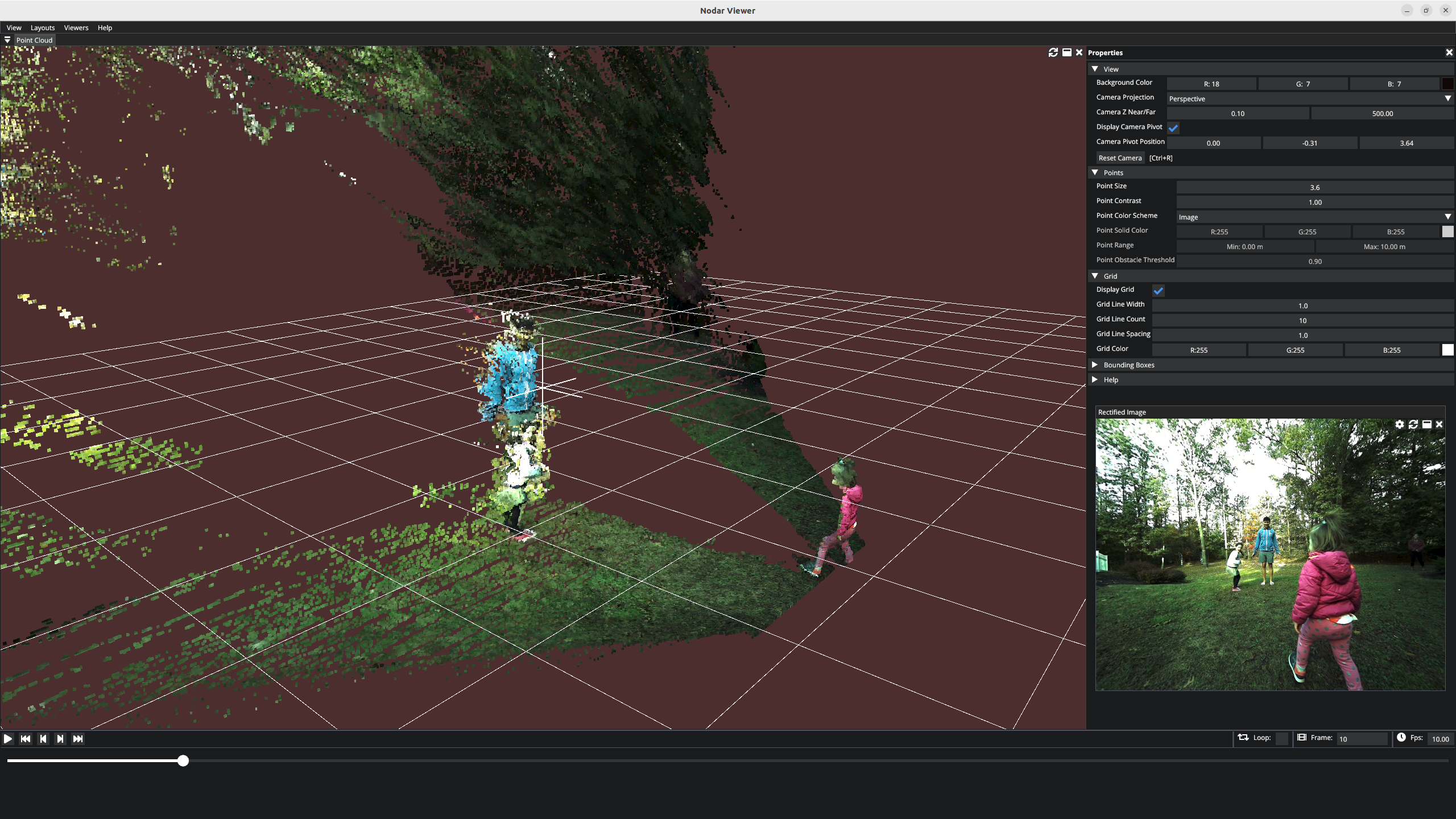}}
\isPreprints{}{
\end{adjustwidth}
} 
\caption{Point cloud from frame 10 from the video: (\textbf{a}) GroundTruth with 5MP images. (\textbf{b}) GroundTruth with 1.3MP images. (\textbf{c}) Hammerhead with 5MP images. (\textbf{d}) Hammerhead with 1.3MP images. Visualizations from NODAR Viewer \cite{viewer}. The grid has 1-m spacing.}
\label{frame10}
\end{figure}

The point clouds for frame 10 are shown in Fig.~\ref{frame10}. The GroundTruth point clouds have fewer stray points than the Hammerhead point clouds, which are mainly seen at the edges of objects, indicating some foreground-background mixing. Not surprisingly, the downsampled images (factor of 2 in rows and columns, or 1280 x 992 pixels) show a loss of point cloud quality and density. Videos of the point clouds can be found here \cite{videos}. The point clouds reveal a gentle downward slope in the grass surface, with about 1 meter of elevation change for every 20 meters of range. Monocular depth estimation networks often struggle to produce metrically accurate depth reconstructions, indicating the importance of direct measurements, such as triangulation (i.e., stereo vision).

\begin{figure}[H]
\isPreprints{}{
\begin{adjustwidth}{-\extralength}{0cm}
\centering
} 
\subfloat[\centering]{\includegraphics[width=7.0cm]{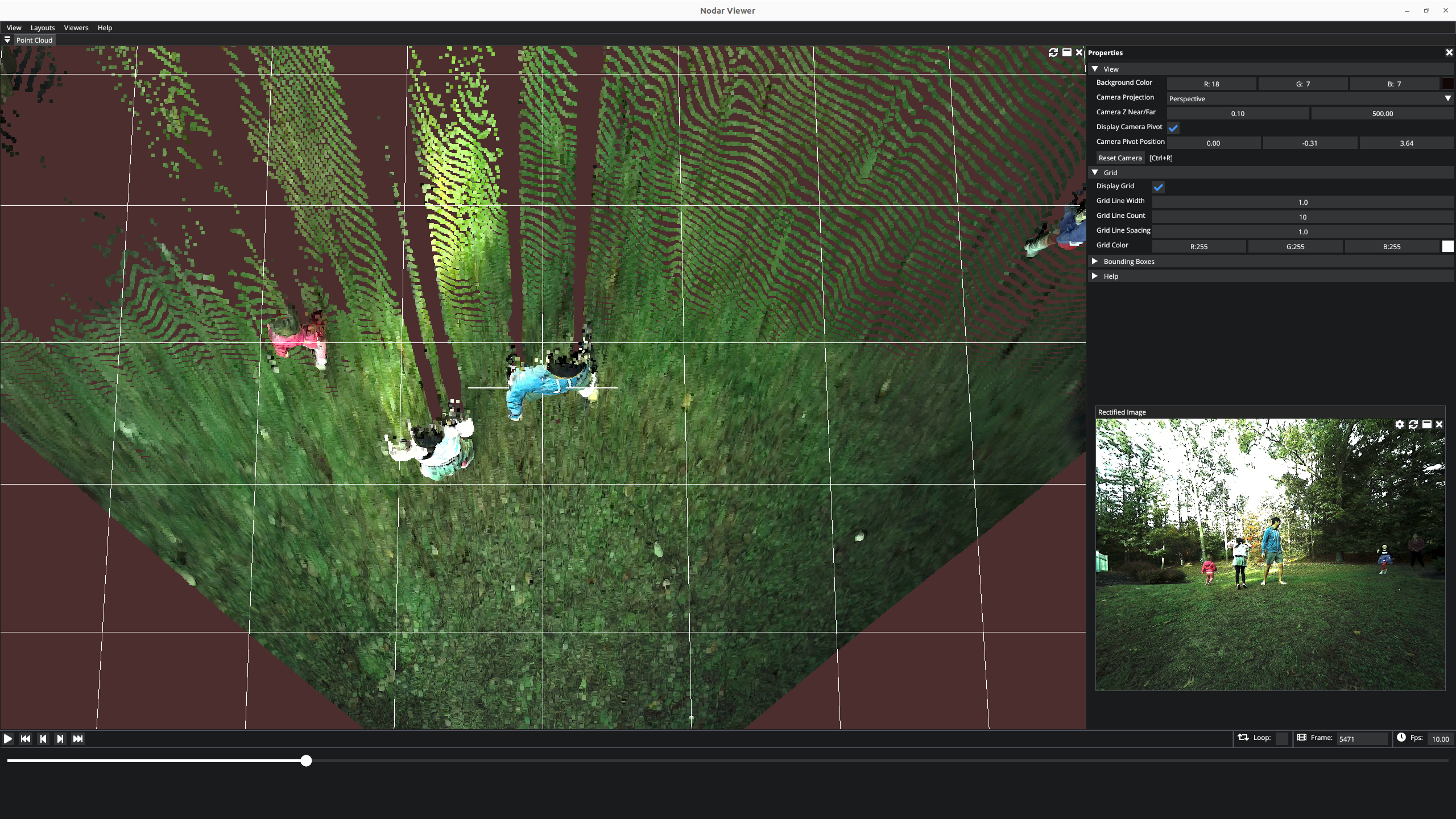}}
\subfloat[\centering]{\includegraphics[width=7.0cm]{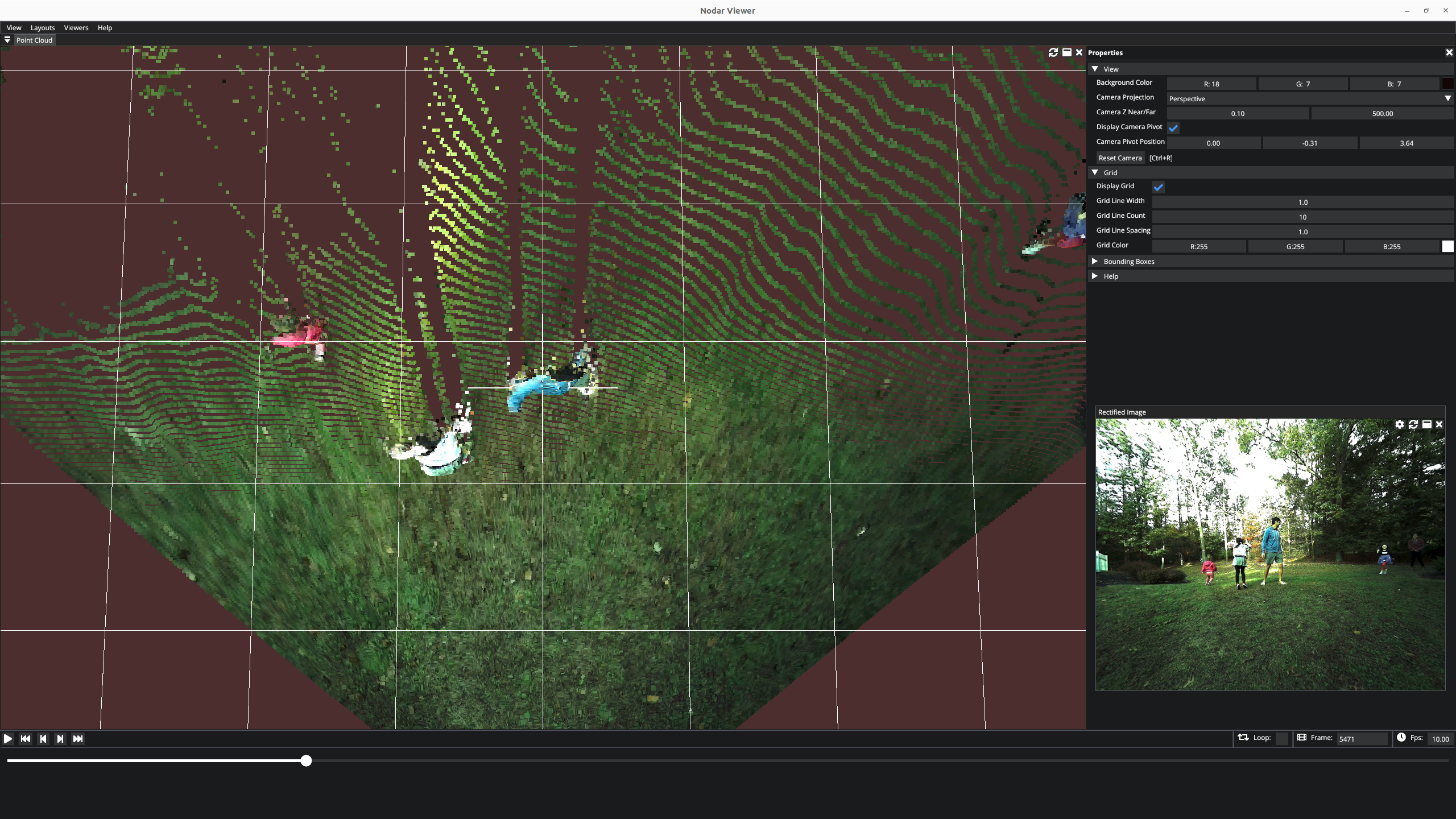}}\\
\subfloat[\centering]{\includegraphics[width=7.0cm]{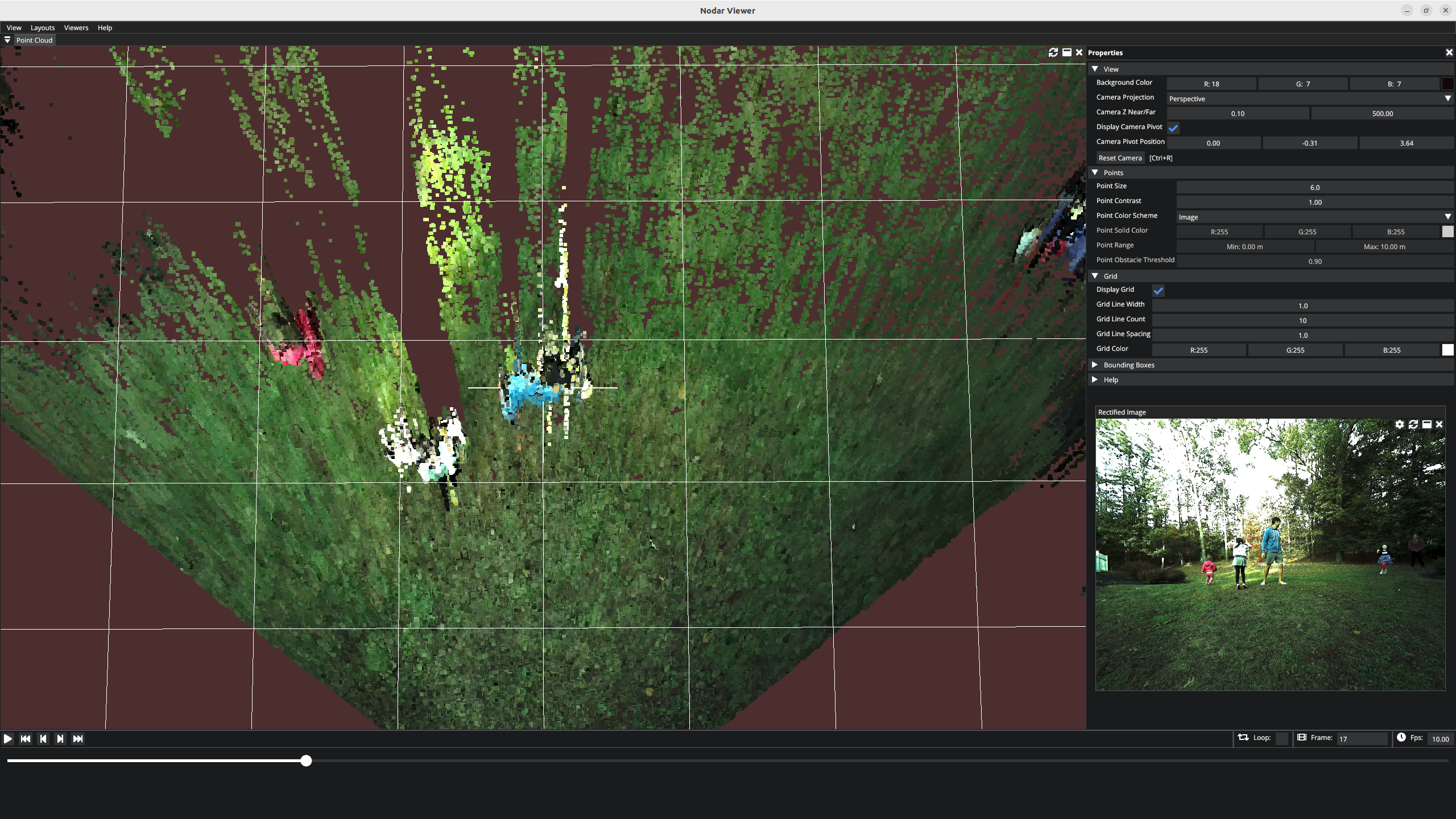}}
\subfloat[\centering]{\includegraphics[width=7.0cm]{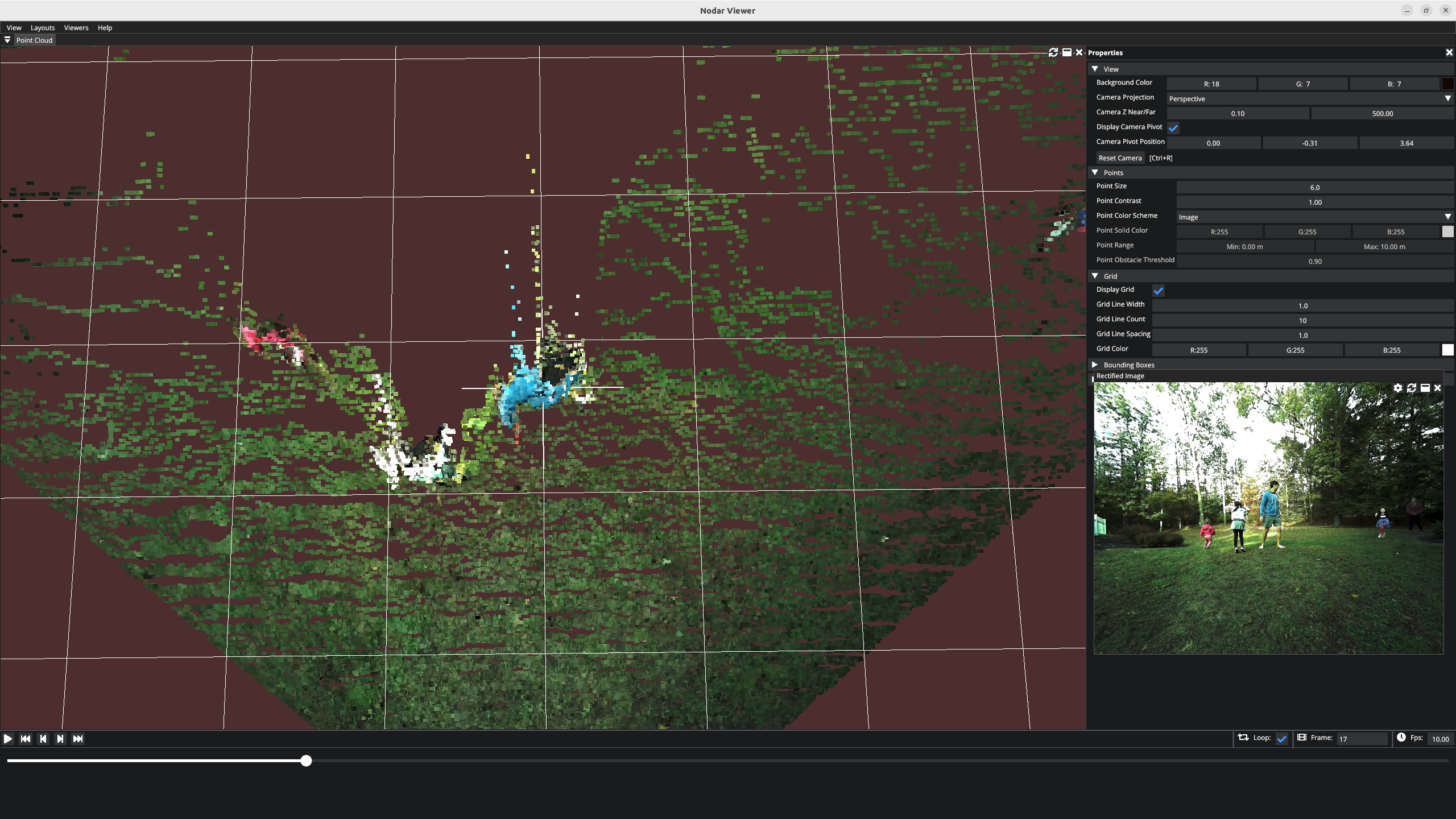}}
\isPreprints{}{
\end{adjustwidth}
} 
\caption{Top view of point cloud from frame 17 from the video centered on an adult in blue shirt: (\textbf{a}) GroundTruth with 5MP images. (\textbf{b}) GroundTruth with 1.3MP images. (\textbf{c}) Hammerhead with 5MP images. (\textbf{d}) Hammerhead with 1.3MP images. Visualizations from NODAR Viewer \cite{viewer}. The grid has 1-m spacing.}
\label{leaf}
\end{figure}

The top view of the point cloud of the adult in the blue shirt at a depth of 3.64 m is shown in Fig.~\ref{leaf} for frame 17. The top view (or Bird's-eye view) is seldom shown in stereo vision publications because it reveals range errors, especially at the edges of objects and at longer ranges. However, even at a relatively long depth of 3.64 m (or 72.8x baseline and wide-FOV lenses), the point cloud quality is high because the larger-format 5MP imager enables more accurate angular measurements.

The point cloud top view (or Bird's-eye view) of an adult wearing a blue shirt is presented in Fig.~\ref{leaf} (for frame 17), captured at a depth of 3.64 m. While the top view is often omitted in stereo vision literature -- as it inherently exposes range errors, particularly at object boundaries and extended ranges -- this visualization clearly demonstrates the system's high point cloud quality. This superior performance, even at a relatively long range (3.64 m, corresponding to $24.3 \times \text{baseline}$ with wide-FOV lenses), is attributed to the use of a larger-format 5MP imager, which facilitates more accurate angular measurements. It's also worth noting that the 5MP point clouds reveal more points between objects or between legs than the lower-resolution 1.3MP depth maps.

\begin{figure}[H]
\isPreprints{}{
\begin{adjustwidth}{-\extralength}{0cm}
\centering
} 
\subfloat[\centering]{\includegraphics[width=7.0cm]{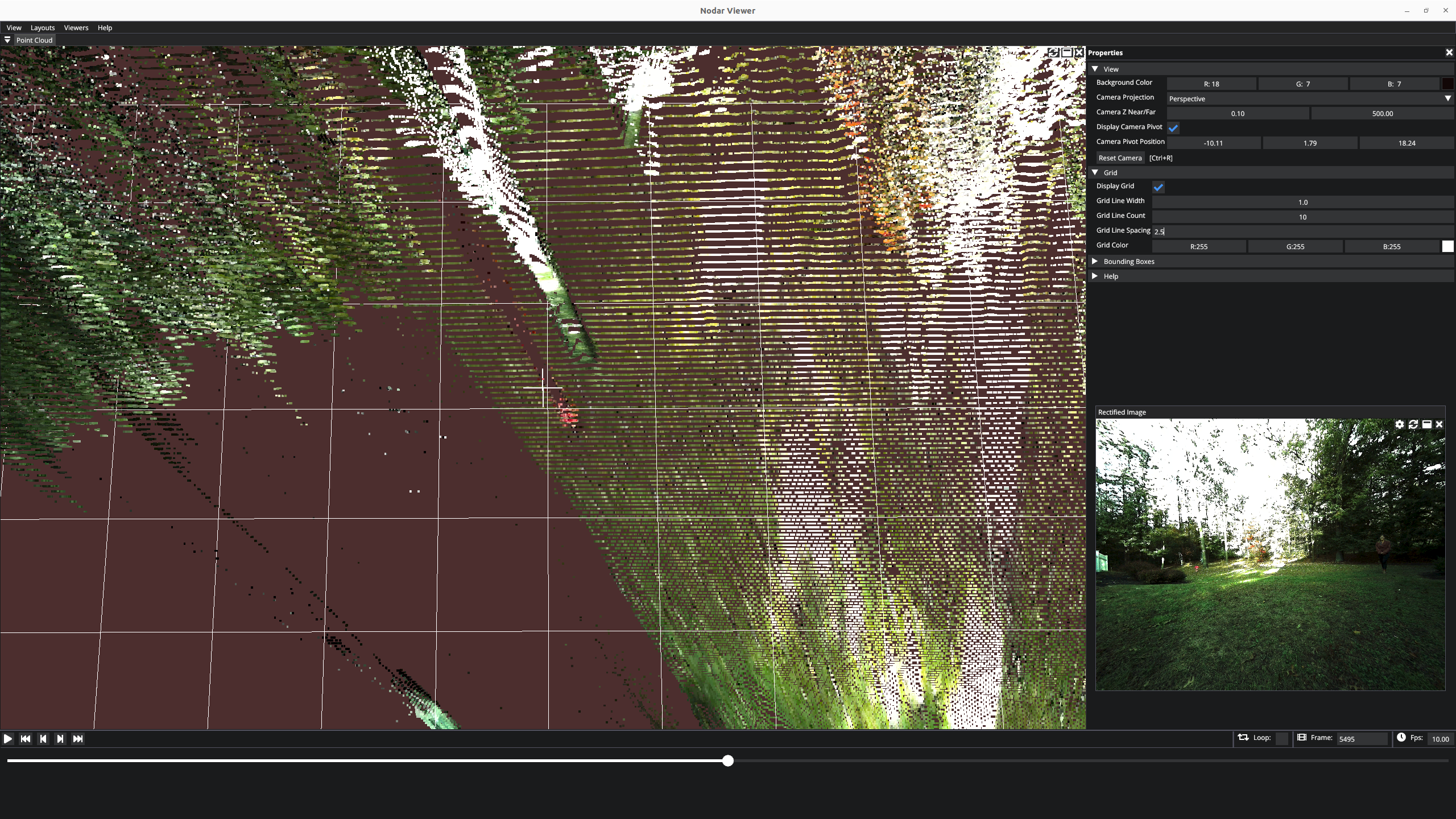}}
\subfloat[\centering]{\includegraphics[width=7.0cm]{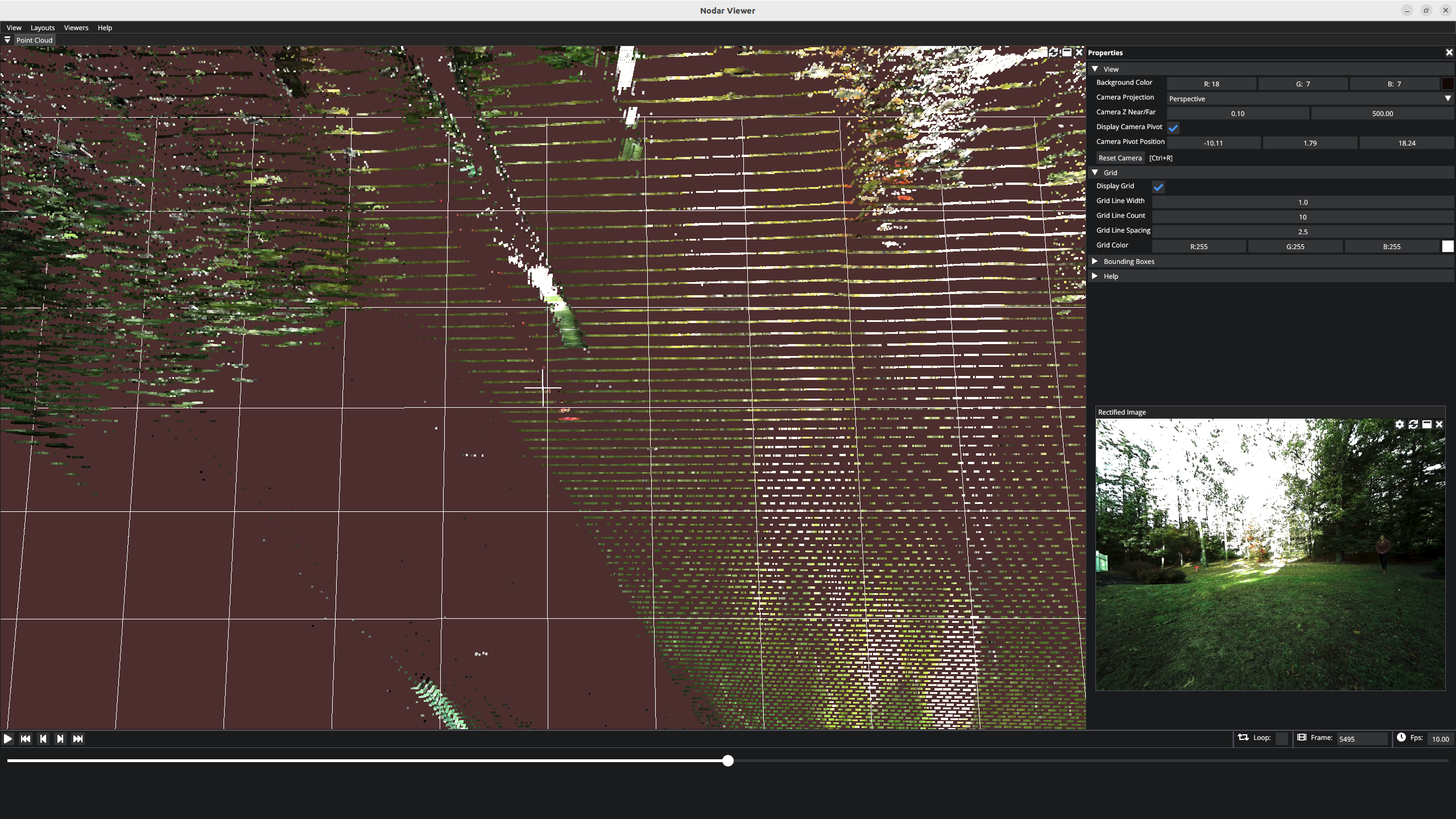}}\\
\subfloat[\centering]{\includegraphics[width=7.0cm]{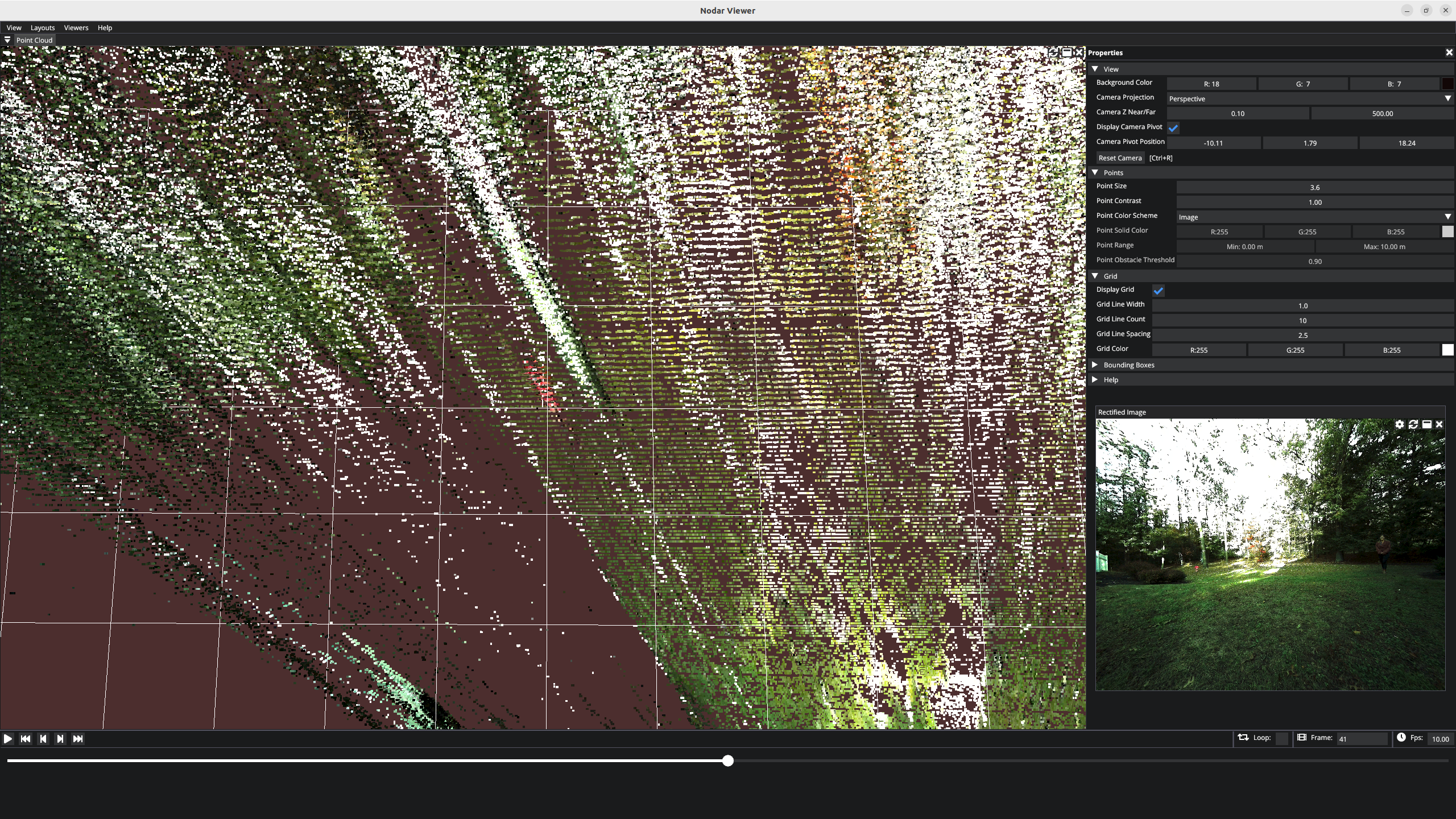}}
\subfloat[\centering]{\includegraphics[width=7.0cm]{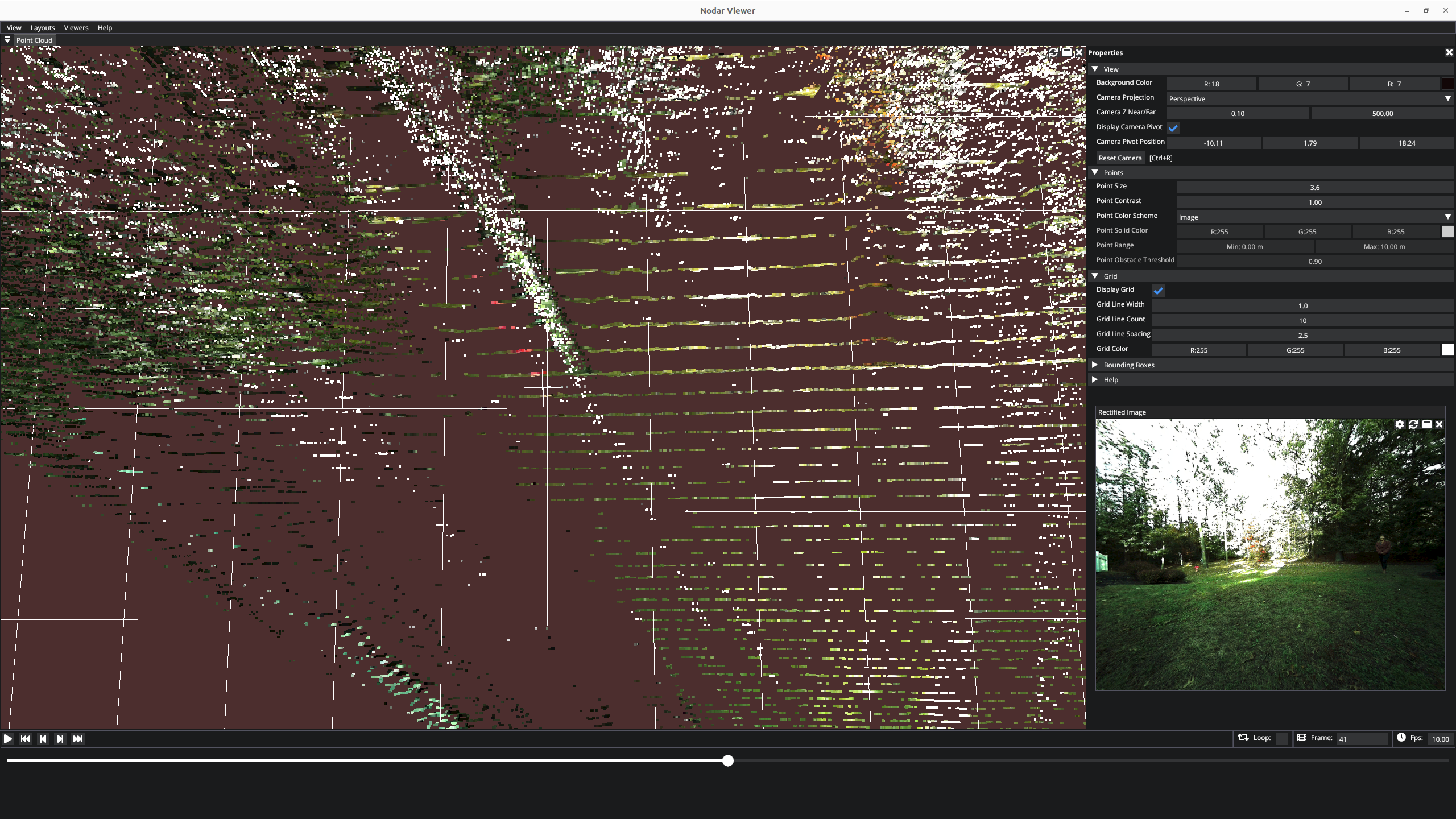}}
\isPreprints{}{
\end{adjustwidth}
} 
\caption{Top view of point cloud from frame 41 from the video centered on a child in a pink jacket next to the tree on the left side of the image. The child is located in the center of the bird's eye view map (pink points). (\textbf{a}) GroundTruth with 5MP images. (\textbf{b}) GroundTruth with 1.3MP images. (\textbf{c}) Hammerhead with 5MP images. (\textbf{d}) Hammerhead with 1.3MP images. Visualizations from NODAR Viewer \cite{viewer}. The grid has 2.5-m spacing.}
\label{callie}
\end{figure}

The point cloud top view (or Bird's-eye view) of a child wearing a pink jacket is presented in Fig.~\ref{callie} (for frame 41), captured at a depth of 18.24 m, corresponding to $121.6 \times \text{baseline}$. This frame corresponds to an extremely long range. At this extreme range, the higher-resolution 5MP point clouds reveal the true advantage to longer ranges: points are clustered closer together and less range spread from objects.

\section{Discussion}

\begin{figure}[H]
\isPreprints{}{
\begin{adjustwidth}{-\extralength}{0cm}
\centering
} 
\subfloat[\centering]{\includegraphics[width=10.0cm]{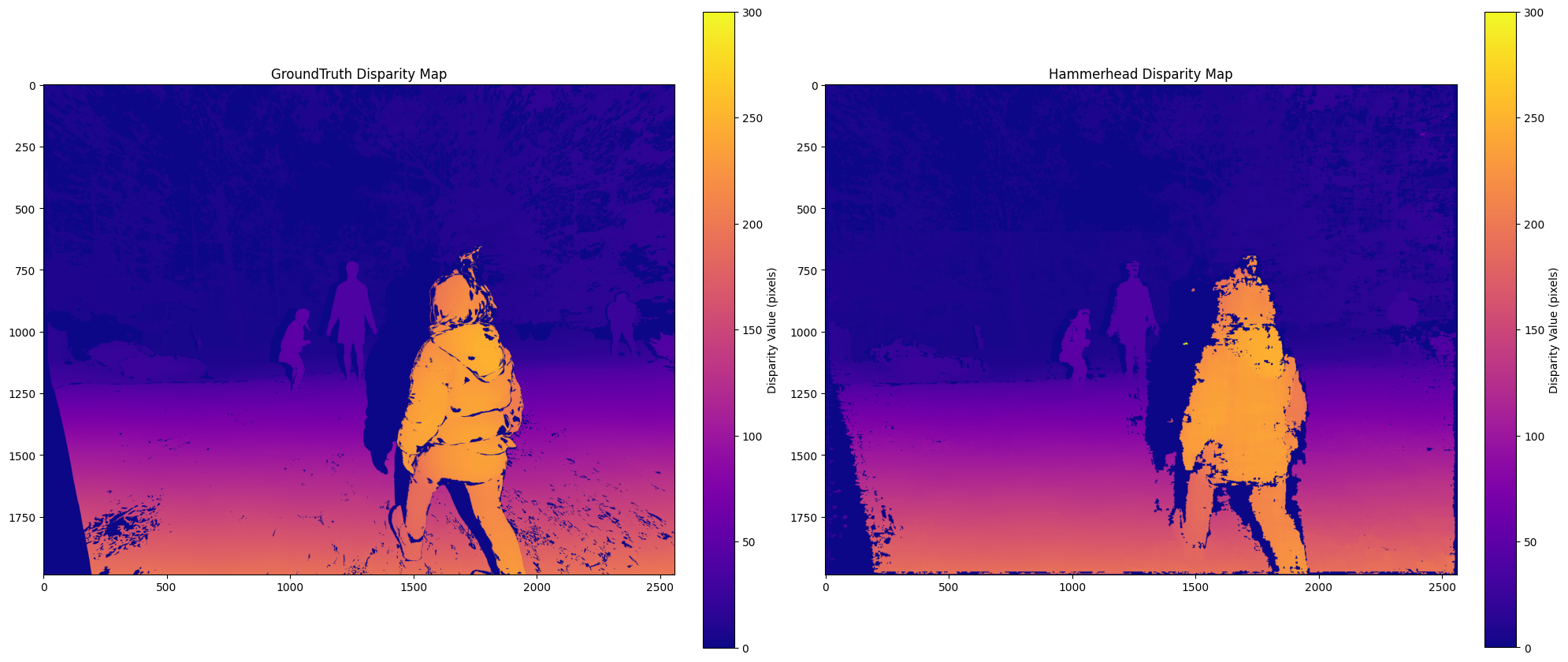}}
\subfloat[\centering]{\includegraphics[width=5.0cm]{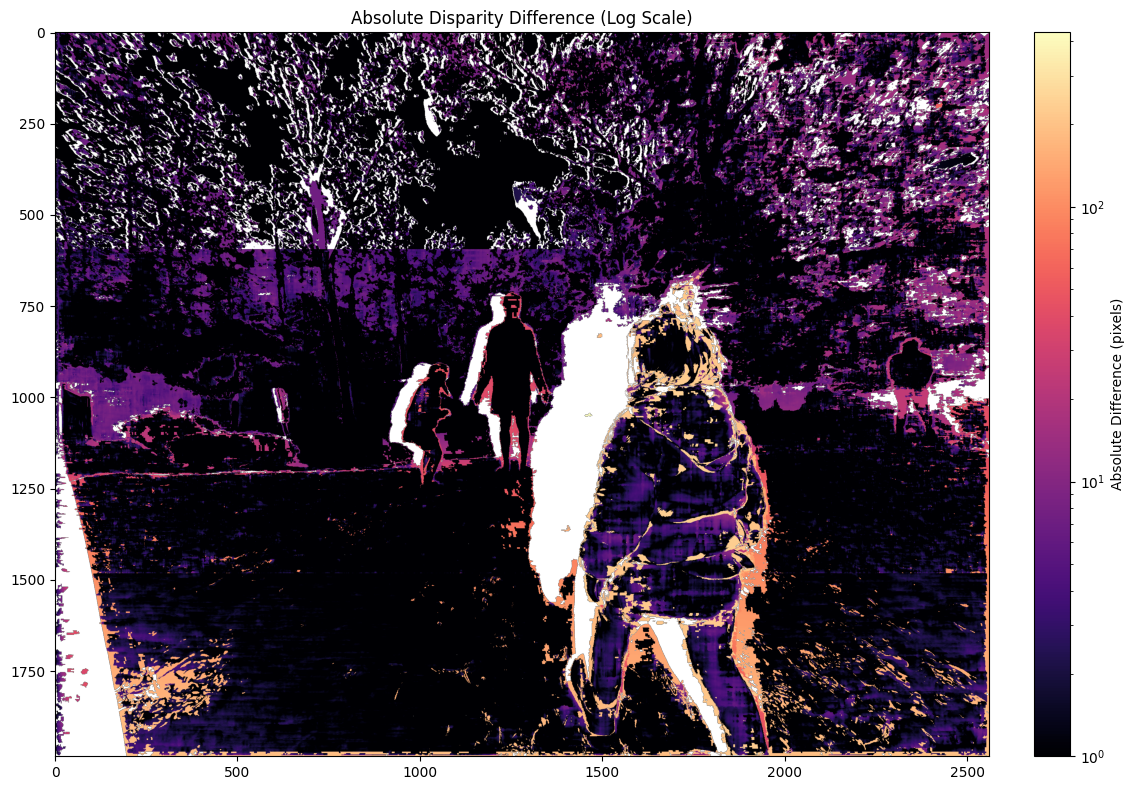}}
\isPreprints{}{
\end{adjustwidth}
} 
\caption{Disparity maps from frame 10 (full 5 MP resolution) from the video: (\textbf{a}) GroundTruth algorithm (left). Hammerhead algorithm (right). (\textbf{b}) The absolute difference between GroundTruth and Hammerhead disparity maps.}
\label{disparity_map}
\end{figure}
\begin{figure}[H]
\isPreprints{\centering}{} 
\includegraphics[width=1.0\textwidth]{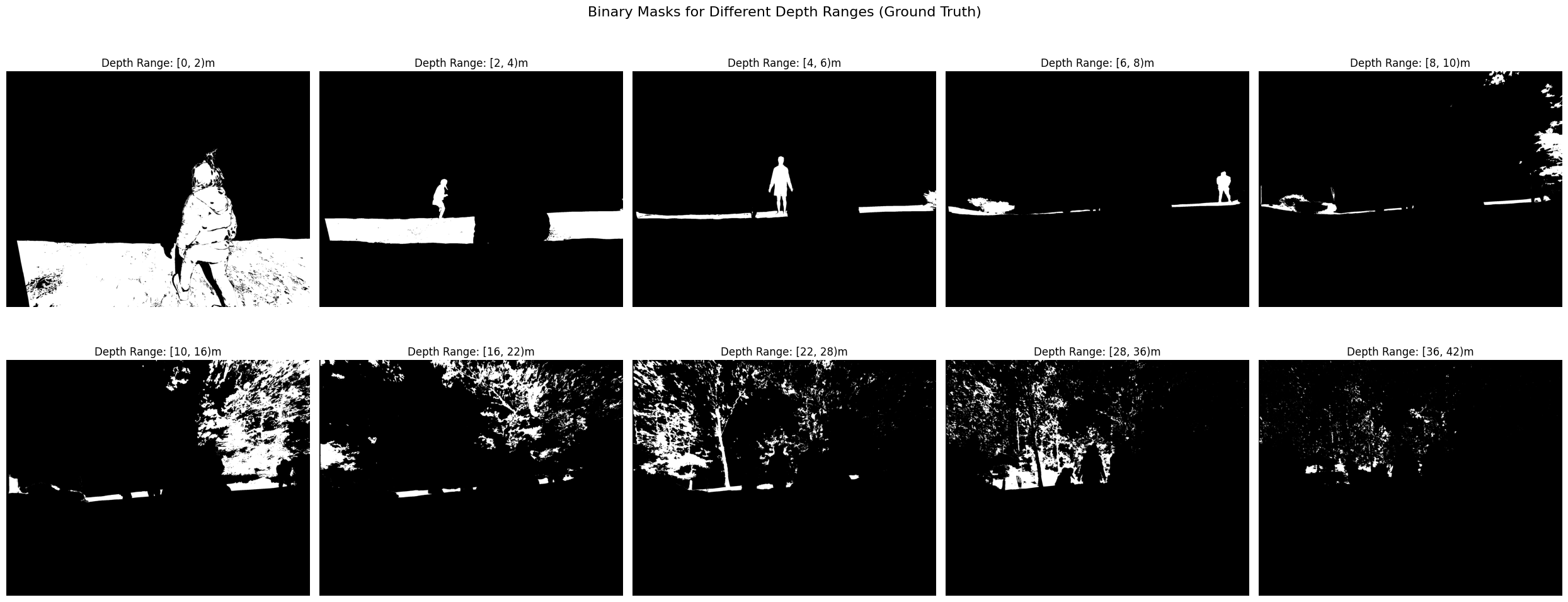}
\caption{Binary masks corresponding to 10 different depth bins for frame 10.}
\label{binary_masks}
\end{figure}

The point cloud performance of the real-time Hammerhead algorithm is evaluated for full and half-resolutions, and for frames 10, 17, and 41 (corresponding to people at close, medium, and far range) as follows:
\begin{enumerate}
    \item The real-time (Hammerhead) and offline (GroundTruth) disparity maps are loaded.
    \item The disparity maps are visualized side-by-side. See Fig.~\ref{disparity_map}(a).
    \item The absolute difference between the two disparity maps is computed and plotted. See Fig.~\ref{disparity_map}(b).
    \item Binary masks based on range bins are computed. Each mask corresponds to the pixels at a given depth bin. The depth bins for this work were (0, 2), (2, 4), (4, 6), (6, 8), (8, 10), (10, 16), (16, 22), (22, 28), (28, 36), and (36, 42) meters. At close ranges, the bins were 2 meters wide up to 10 meters depth, and 6 meters wide up to 42 meters depth. The wider bins at longer ranges help account for the point spread.
    \item Iterate through the binary masks, apply each mask to the absolute disparity difference map, and compute the median absolute disparity error (in meters). Plot the measured results against a theoretical depth error curve (square of depth, $Z^2$). See Fig.~\ref{binary_masks}.
\end{enumerate}
The source code for computing the processed results is available in a publicly accessible Google Colab notebook \cite{colab}. The GroundTruth and Hammerhead disparity maps (e.g., Fig.~\ref{disparity_map}(a)) are very similar with larger differences at the edges of the object since Hammerhead's algorithm slightly overestimates the size of objects by a few pixels (see Fig.~\ref{disparity_map}(b)). 

The GroundTruth disparity map is masked into 2-m depth bins from $Z = 0$ to 10 meters and into 6-m depth bins from $Z=10$ to 42 meters. Making the depth bins larger at greater ranges helps accumulate more statistics for estimating depth error, as the fraction of the image corresponding to longer ranges decreases. For example, the binary masks corresponding to the depth bins is shown in Fig.~\ref{binary_masks}, which shows a statistically significant number of pixels in each depth bin.

\begin{figure}[H]
\isPreprints{\centering}{} 
\includegraphics[width=0.75\textwidth]{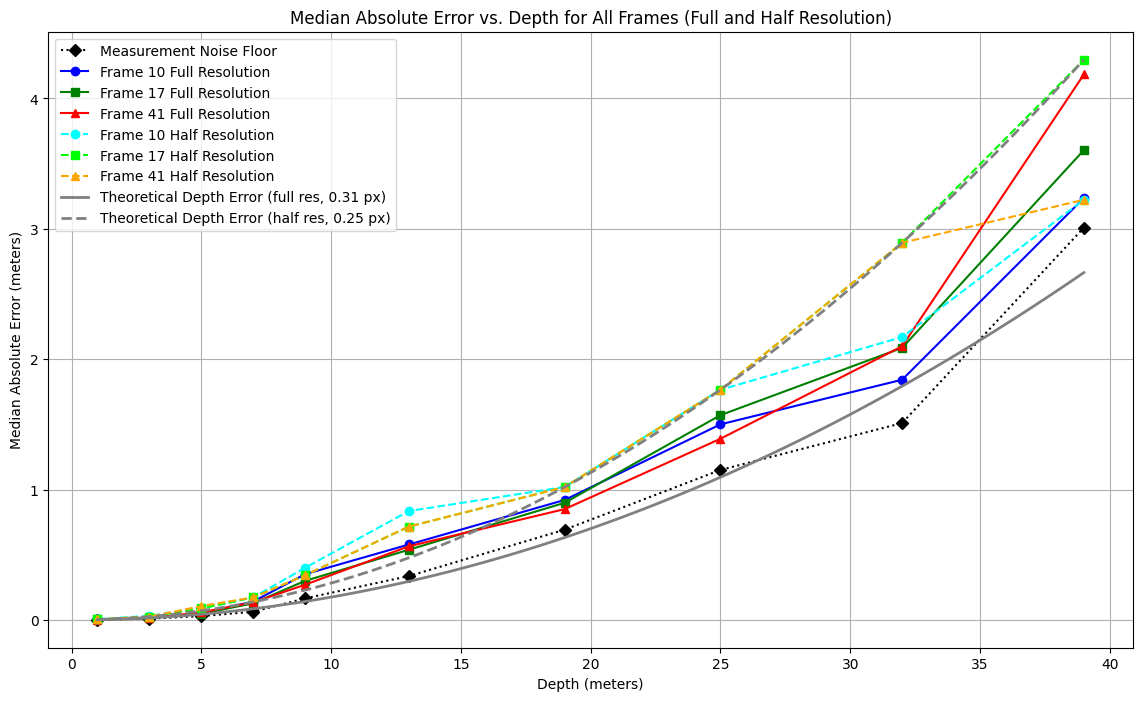}
\caption{The mean absolute error (meters) vs. depth.}
\label{median_abs_error}
\end{figure}

The primary results of this study are presented in Fig.~\ref{median_abs_error}, which illustrates the relationship between mean absolute error and depth. As expected, depth error increases with the object's distance from the stereo sensor. To calculate this error, we determined the median absolute difference between the Hammerhead and Ground Truth disparity maps, subsequently converting the values from pixels to meters. The median was utilized rather than the mean to ensure the results remain robust against outliers.

The noise floor of the measurement (black dotted line with diamonds) is computed for frame 10, by taking the difference of the GroundTruth disparity map of full-resolution images (followed by downsampling) and the GroundTruth disparity map of half-resolution images. As expected, the measurement noise floor is below all other experimental data.

The theoretical depth errors ($\Delta Z = Z^2/(fB) \Delta d$) are plotted for full resolution images ($f = 1180$ pixels and $\Delta d = 0.31$  pixels, solid gray line), and half resolution images ($f = 590$ pixels and $\Delta d = 0.25$ pixels, dashed gray line). The pixel error ($\Delta d$) was tuned to fit the measured data. As expected, the full-resolution depth maps offer better depth measurement accuracy than the half-resolution depth maps. For example, at a depth of $Z=39$ meters, the mean absolute errors are 2.6 and 4.3 meters, respectively.

The median absolute errors vs. depth for frames 10, 17, and 41 are shown with solid lines for full-resolution images (5 MP) and dashed lines for half-resolution images. In general, the error for full-resolution depth maps is better than that of half-resolution depth maps.

The mean absolute error of points is sub-meter for objects up to around 20-meter distance, which is a direct result of processing high-resolution 5 MP images in real-time.

\begin{figure}[H]
\isPreprints{\centering}{} 
\includegraphics[width=0.75\textwidth]{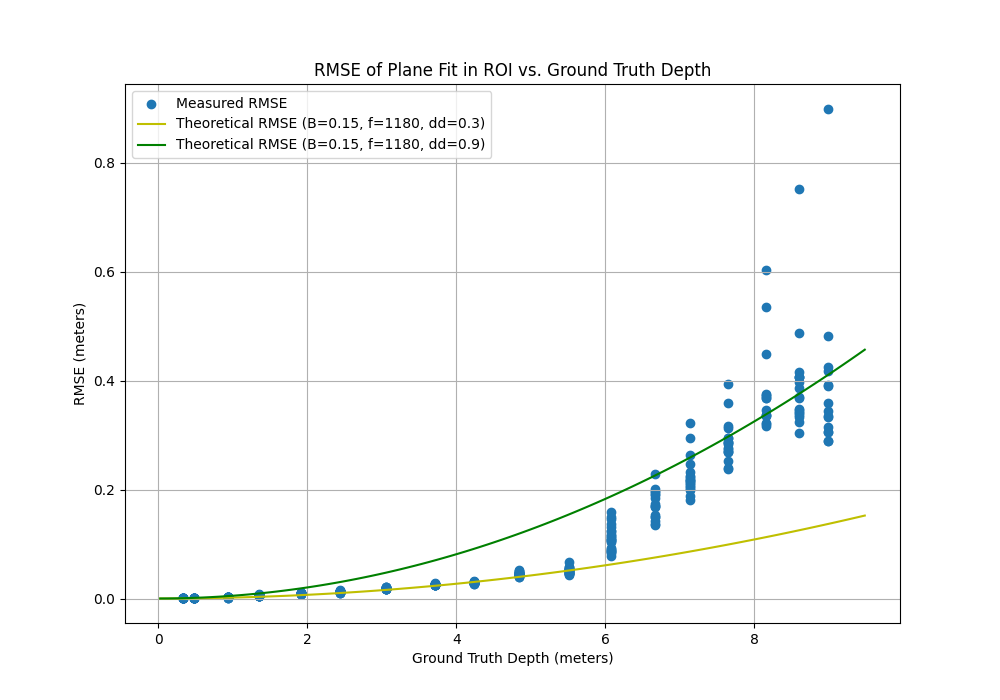}
\caption{The RMSE (meters) vs. depth.}
\label{rmse}
\end{figure}

To verify these results, a more traditional approach for measuring the error was to use a flat board with a textured pattern. A plane was fit to the depth map over the region of interest (ROI) corresponding to the flat board, at ranges from 0.33 to 8.98 meters, in approximately half-meter increments. The root-mean-square error (RMSE) between the fitted plane and the depth map values was then computed over the 100 x 100 pixel ROI. The plot of RMSE vs. depth is shown in Fig.~\ref{rmse}. The corresponding error agrees quite well with Fig.~\ref{median_abs_error}.

\section{Conclusions}

As stereo vision cameras increase in resolution, they are able to support detection at longer ranges due to inherently better angular measurements. Increasing the resolution of stereo cameras by a factor of 5 (e.g., 1 MP to 5 MP), increases the range of the sensor by a factor of 2.24. We presented a technique for quantifying the depth map quality based on the GroundTruth algorithm. As the angular resolution and baseline of the cameras increase, calibration becomes a necessary condition.

Under typical conditions, a wide-angle stereo camera with a 137-degree field of view and only 15-cm baseline is not expected to reliably detect objects at a distance of 20 m. However, the increased spatial resolution of our system enables effective perception at this range.



\vspace{6pt}

\informedconsent{Informed consent was obtained from all subjects involved in the study.}


\dataavailability{The original data presented in the study are openly available in \cite{videos}, \cite{gdrive_data}, and \cite{nodarhub_leopard}.}

\acknowledgments{The authors thank Leopard Imaging for providing the Eagle stereo vision camera used in the study.}


\conflictsofinterest{The authors declare no conflicts of interest.}

\isPreprints{}{
\begin{adjustwidth}{-\extralength}{0cm}
} 

\reftitle{References}


\bibliography{refs}

\isPreprints{}{
\end{adjustwidth}
} 
\end{document}